\documentclass{article}
\usepackage{spconf,amsmath,graphicx}

\usepackage{cite}
\usepackage{amsmath,amssymb,amsfonts}
\usepackage{algorithmic}
\usepackage{graphicx}
\usepackage{textcomp}
\usepackage{hyperref}
\usepackage{booktabs, times, epsfig, graphicx, amsmath, amssymb,url, multirow,comment,commath}

\graphicspath{ {models/} }

\def\equationautorefname~#1\null{(#1)\null}

\title{Multi Label Restricted Boltzmann Machine  for Non-Intrusive Load Monitoring}
\name{Sagar Verma, Shikha Singh and Angshul Majumdar}
\address{Indrapratha Institute of Information Technology, Delhi}
\begin{document}
%
\maketitle

\maketitle
\thispagestyle{empty}
\pagestyle{empty}

\begin{abstract}
Increasing population indicates that energy demands need to be managed in the residential sector. Prior studies have reflected that the customers tend to reduce a significant amount of energy consumption if they are provided with appliance-level feedback. This observation has increased the relevance of load monitoring in today's tech-savvy world.  Most of the previously proposed solutions claim to perform load monitoring without intrusion, but they are not completely non-intrusive. These methods require historical appliance-level data for training the model for each of the devices. This data is gathered by putting a sensor on each of the appliances present in the home which causes intrusion in the building. Some recent studies have proposed that if we frame Non-Intrusive Load Monitoring (NILM) as a multi-label classification problem, the need for appliance-level data can be avoided. In this paper, we propose Multi-label Restricted Boltzmann Machine(ML-RBM) for NILM and report an experimental evaluation of proposed and state-of-the-art techniques.

\end{abstract}


\section{INTRODUCTION}
Increasing population is a direct indication of growth in energy demands in the residential as well as commercial sector \cite{energy_use}.  To meet the demands non-renewable resources are used which is the leading cause of global warming\cite{archer2012global}. Energy saving and demand-side management is need-of-the-hour.

Load monitoring is one of the ways to collect the energy data required to devise automated energy management systems.  It also helps in providing feedback to consumers to understand their consumptions. The feedback enables customers to participate in energy-saving and cost-cutting activities\cite{darby2006effectiveness}.

There are two ways to perform load monitoring. One way is to put sensors on each of the appliances deployed in a building that can sample the energy consumption and store the data. The second way is to use such algorithms that can segregate the appliance-level load from the aggregated units of consumption. The first one is a non-practical approach. It is not only costly but intrusive too. The second way, known as Non-Intrusive Load Monitoring /Non-Intrusive Appliance Load Monitoring (NIALM), is more practical and scalable than the first one.

Most of the state-of-the-art NILM algorithms, until recently, use historical appliance-level sampled data to learn models of individual appliances used in the building. Once the model for each of the targeted devices is trained, the segregation could be performed just by estimating the device-specific loads from the sampled aggregated load. The requirement of appliances' consumption data makes the training phase intrusive.

In recent studies \cite{tabatabaei2017toward,li2016whole, singhal2018simultaneous}, NILM has been framed as a multi-label classification problem.  These techniques make use of annotated aggregated load for training the model.  The annotation contains information about the ON/OFF state of each of the target devices. The annotation can be performed with the help of the cameras installed on premises. This framework circumvents the need for device-level loads, and thus the training phase does not require multiple sensors in the buildings. This modification enables a large-scale roll-out of NILM from the utilities.

Our work is motivated by advantages of the transformation of NILM as a Multi-label classification task and the success of deep learning as a solution to similar problems. We propose a new approach to multi-label classification based on the Restricted Boltzmann Machine (RBM)\cite{Larochelle2008CUD}. RBMs have never been used for multi-label classification so far. It is a classic example of algorithm adaptation for multi-label classification.

RBMs \cite{Smolensky1986} have been effective in learning high-level features and capturing high-order correlations of the observed variables. A typical RBM has a hidden unit in which nodes are conditionally independent given the visible state. RBMs have good reconstruction accuracy which can be leveraged to generate individual load information in latent space. We propose that generative property of RBMs combined with multi-label supervision can be used to perform NILM via state detection of appliances.

\section{Literature review}

\subsection{\textit{ Combinatorial Optimization }}

Studies in combinatorial optimization (CO) such as \cite{hart1992nonintrusive} are based on the principle that total consumption in a building can be approximated as a sum of device-level loads. So aggregated consumption in a building can be expressed as

\begin{equation}
    {P_{agg}} = \sum\limits_{i = 1}^N {{s_i}{P_i}}
\end{equation}

 where \textit{$P_i$} is individual device load, \textit{$P_{agg}$} is aggregated load,  \textit{N} is the total number of appliances,  and \textit{$s_i$} is a vector that indicates the state of devices i.e., 0 for 'OFF' state and 1 for 'ON' state.

For load segregation, the motive is to find out the combinations of individual loads whose sum can be approximated as the aggregated load. We can formulate the task of simultaneous detection of ON/OFF state of the devices, $\hat s$, as

\begin{equation}
 \mathord{\buildrel{\lower3pt\hbox{$\scriptscriptstyle\frown$}}
\over s}  = \mathop {\arg \min }\limits_s \left| {P_{agg} - \sum\limits_{i = 1}^N {{s_i}{P_i}} } \right|
\end{equation}

 Equation (2) is an NP-hard problem and quickly becomes intractable as the number of appliances scales up.

\subsection{\textit{ Finite State Machines }}
Apart from the computational complexity, another problem with CO is that it cannot account for the fact that one appliance can run at different power levels, e.g. A.C., fan, washer, etc. However these days most of the appliances (like light, fan, A.C., washer) have marked different states, so it is fair to model the state of the devices as Hidden Markov Models (HMMs). The study \cite{kim2011unsupervised} models aggregated load as an outcome of the interaction of a finite number of independent HMMs.

 Most of the modern appliances such as printers, computers, inverters do not have marked states. They are continuously varying. In such situations, the above assumption fails; this, in turn, leads to poor disaggregation performance.

\subsection{\textit{ Multi-Label Classification }}

The classification task where one sample may belong to one or more classes is known as multi-label classification (MLC). Hence, in this case, each sample is mapped to a binary vector of 0's and 1's, assigning 0 or 1 to each of the classes.

Since the aggregated load of a building at any instance may be an outcome of several active appliances' consumption, Tabatabaei et al. \cite{tabatabaei2017toward}, and Li et al. \cite{li2016whole}, framed NILM as an MLC problem. \cite{tabatabaei2017toward} compared the performance of two multi-label classifiers viz Multi-Label K-Nearest Neighbours (ML-kNN) and Random k-Label Sets (RakEl) using time-domain and wavelet-domain features of appliances.

Another recent work \cite{singhal2018simultaneous} uses Multi-label Consistent Deep Dictionary Learning for simultaneous detection of active appliances from smart meter data. These methods do not directly segregate appliance-level load but first identify states of appliances and then disaggregated load is obtained by multiplying the average power consumption of device with the number of instances, it was detected to be in an active state. By far these are the most recent and best-known techniques for multi-label classification based disaggregation.

\section{Proposed approach}
Restricted Boltzmann Machines \cite{Smolensky1986} are one type of undirected graphical models that use hidden variables to model high-order and non-linear regularities of the data. A typical RBM is a two-layer bipartite graph with two types of units, the visible units $x$ and hidden units $h$. An RBM represents probability distributions over the random variables under an energy-based model. The energy model of an RBM is given by $E(x,h) = -x^TWh-b^Tx-c^Th$, where $W$ is the weight to be learned. The joint probability distribution over $(x,h)$ is expressed as $P(x,h) = \frac{1}{z}exp(-E(x,h))$, where $Z$ is the normalization factor. Learning RBMs is a difficult task due to the tractability involved in computing normalization factor $Z$. Several learning algorithms have been proposed \cite{CD2002hinton, Larochelle2012, pmlr-v9-marlin10a} to solve the problem above. Contrastive Divergence (CD) method proposed by Hinton et al. \cite{CD2002hinton} is an efficient method and is widely used to learn RBMs. The generative property of RBM makes it useful for learning latent space representation of data where we don't have information about how data is generated. RBMs have been used for dimensionality reduction \cite{Hinton504}, collaborative filtering \cite{Salakhutdinov2007RBM}, anomaly detection \cite{FIORE2013anomaly} and unsupervised feature learning \cite{pmlr-v15-coates11a}. The classification RBM has been used for various classification tasks in \cite{Larochelle2008CUD, Li2015ConditionalRB} and label consistent collaborative filtering \cite{Verma2018CollaborativeFW}.

\subsection{Multi-Label Classification RBM}

The joint probability distribution of the proposed multi-label classification RBM model shown in figure \ref{fig:1} is given by,

\begin{equation}
  p(y,x,h) \propto e^{-E(y,x,h)}
  \label{eq:1}
\end{equation}

where $y$ is the label unit. We define the new energy function as follows:

\begin{figure}[!t]
  \centering
  \includegraphics[width=0.8\linewidth]{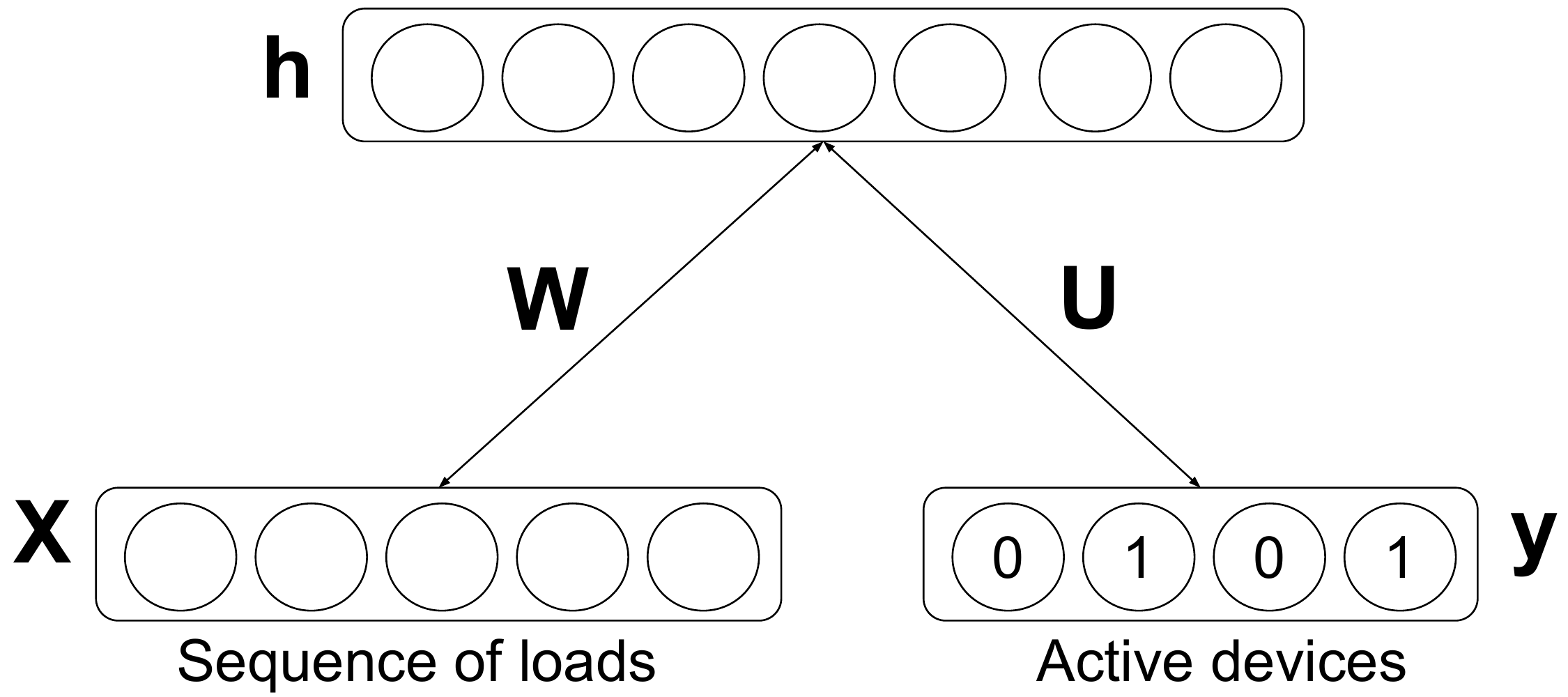}
  \caption{Proposed architecture for NILM using multi-label classification RBM.}
  \label{fig:1}
 \end{figure}

\begin{equation}
  E(y,x,h) = -h^TWx - a^Tx -b^Th - c^Ty - h^TUy
  \label{eq:2}
\end{equation}

with parameters $\Theta = (W,a,b,c,U)$. The model is illustrated in figure \ref{fig:1}. We find the values of visible and hidden units using \autoref{eq:3}, \autoref{eq:4} and \autoref{eq:5} respectively.

\begin{equation}
  p(h_j=1|x,y) = \sigma(b_j+U_{jl}+\sum_kW_{ji}x_i)
  \label{eq:3}
\end{equation}

\begin{equation}
  p(x_i|h) = \sigma(a_i + \sum_jW_{ji}h_j)
  \label{eq:4}
\end{equation}

\begin{equation}
  p(y_l=1|h) = \frac{exp(c_l+\sum_{j}U_{jl}h_j)}{\sum_{l=1}^{y}exp(c_l+\sum_{j}U_{jl}h_j)}
  \label{eq:5}
\end{equation}

where $\sigma$ is the logistic sigmoid and $l$ is the class label out of $C$ classes. These formulations capture the predictive information about the input vector as well as the target class.

Network parameter $\Theta$ is learned using CD \cite{CD2002hinton} algorithm,

\begin{align}
  \Delta W_{ij} & = \eta \frac{\delta logp(x,y)}{\delta W_{ij}} \nonumber \\
                  & = \eta (<x_iy_ih>_{data}-<x_iy_ih>_{model})
                  \label{eq:6}
\end{align}
where $\eta$ is the learning rate.

For multi-label classification RBM, the above formulation changes as now we have multi-label information present for each sample. The conditional distribution of $y$ given $h$ becomes:

\begin{equation}
  p(y_{l}=1|h) = \sigma(c_l + \sum_iU_{jl}h_j)
  \label{eq:7}
\end{equation}

This formulation is not tractable since $y$ now has $2^C$ possible values. Therefor for inference we use mean field (MF) message-passing method for an approximate inference. The MF approach tries to approximate the joint posterior $p(y,h|x)$ by a factorial distribution $q(y,h) = \prod^C_{l=1}\mu^{y_l}_l(1-\mu_l)^{1-y_l}\prod^n_{j=1}\tau^{h_j}_j(1-\tau_j)^{1-h_j}$ that minimizes the Kullback-Leibler (KL) divergence with the true posterior. Running the following message passing procedure to convergence

\begin{align}
    \mu_l & \leftarrow \sigma(c_l + \sum_jU_{jl}\tau_j) \quad \forall l \in \{1,...,C\}, \\
    \tau_j & \leftarrow \sigma(b_j + \sum_bU_{jl}\mu_l+\sum_iW_{ji}x_i) \quad \forall j \in \{1,...,n\}
\end{align}

we can reach a saddle point of the KL divergence, where $\mu_l$ serves as the estimate for $p(y_l=1|x)$ and $\tau_j$ can be used to estimate $p(h_j=1|x)$.

\section{Experimental Evaluation}

\begin{table*}[!ht]
\centering
\caption{\textbf{Appliance-Level Evaluation on REDD } }
\label{tab1}
\begin{tabular}{l cc cc cc cc}
\toprule[0.2mm]
\multirow{2}{0.5cm}{\textbf{Device}}&\multicolumn{2}{c}{\textbf{MLkNN}}&\multicolumn{2}{c}{\textbf{RAkEL}}&\multicolumn{2}{c}{\textbf{LC-DDL}}&\multicolumn{2}{c}{\textbf{MLC-RBM}} \\
 &F1-Score &Error&F1-Score&Error &F1-Score&Error&F1-Score&Error\\
\midrule
Lighting       &0.6476 &0.3718     &0.6760 &0.8213      &0.6216 &0.2608      &\textbf{ 0.6947}& \textbf{0.1762}             \\

Kitchen        &0.5081 &0.4304     &0.6108 &0.6995               &0.6411 &0.3326      &\textbf{0.7213 }& \textbf{0.1273} \\

Refrigerator   &0.5292 &0.3628     &0.6724 &0.5132               &0.6118 &0.2528      &\textbf{0.7186 }& \textbf{0.1644}\\

Washer Dryer   &0.3903 &0.3122     &0.5267 &0.6990               &0.4977 &0.3149      &\textbf{0.6983} & \textbf{0.1963}\\
\bottomrule[0.2mm]
\end{tabular}
\end{table*}

\begin{table*}[!ht]
\centering
\caption{\textbf{Appliance-Level Evaluation on Pecan Street} }
\label{tab2}
\begin{tabular}{l cc cc cc cc}
\toprule[0.2mm]

\multirow{2}{0.5cm}{\textbf{Device}}&\multicolumn{2}{c}{\textbf{MLkNN}}&\multicolumn{2}{c}{\textbf{RAkEL}}&\multicolumn{2}{c}{\textbf{LC-DDL}}&\multicolumn{2}{c}{\textbf{MLC-RBM}} \\
 &F1-Score &Error&F1-Score&Error &F1-Score&Error&F1-Score&Error\\
\midrule
Air Conditioner  &0.6391  &0.1720    &0.6521&0.8565      &0.5882 &\textbf{0.1051}      &\textbf{0.7023} & 0.2334             \\

Dishwasher       &0.6546  &0.1690    &0.6728&0.8490               &0.4871 &0.1501      &\textbf{0.7269} &\textbf{0.1341}    \\

Furnace          &0.6123  &0.1341    &0.6231&0.8415               &0.5572 &\textbf{0.0794}      &\textbf{0.7113}&0.2224    \\

Microwave        &0.5916  &\textbf{0.0727}    &0.6819&0.7301               &0.5533 &0.0795      &\textbf{0.6981}&0.1985    \\
\bottomrule[0.2mm]
\end{tabular}
\end{table*}

We performed the experiments on two standard datasets viz. The Reference Energy Disaggregation Dataset (REDD) \cite{kolter2011redd} and a subset of Dataport dataset \cite{dp} (also known as Pecan Street Dataset) available in non-intrusive load monitoring toolkit (NILMTK) format\cite{batra2014nilmtk} .

The REDD dataset is a moderate size publicly available dataset for electricity disaggregation. The dataset consists of
power consumption signals from six different houses, where for each house, the whole electricity consumption, as well as
electricity consumptions of about twenty different devices are recorded at every second.

The Dataport dataset contains 1-minute circuit level and building level electricity data from 240 houses. It contains per minute readings from 18 different devices: air conditioner, kitchen appliances, electric vehicle, and electric hot tub heater, electric water heating appliance, dishwasher, spin dryer, freezer, furnace, microwave, oven, electric pool heater, refrigerator, sockets, electric stove, waste disposal unit, security alarm and washer dryer.

 We compare our results with multi-label classification algorithm proposed so far for NILM viz. ML-kNN, RakEl, and LC-DDL. Both the datasets were split into training, testing and cross-validation set in a ratio of 50:30:20  respectively. Cross-validation set was used to decide the values of hyper-parameters. We have munged the data such that each sample contains per hour aggregated consumption and corresponding device labels.

\begin{table}[!ht]
\centering
\caption {\textbf{Performance Evaluation on REDD}}
\label{tab3}
\begin{tabular}{c c c c}
\toprule[0.2mm]

     \textbf{Method}    & \textbf{Macro F1-Score} &\textbf{ Micro F1-Score} \\
    \midrule
     MLkNN              &  0.6086          &  0.6143                     \\

     RAkEL              &  0.6290          &  0.6294                       \\

     LC-DDL             &  0.5222          &  0.5262                      \\

     MLC-RBM        &\textbf{0.7082}   &\textbf{0.7157}       \\
    \bottomrule[0.2mm]

\end{tabular}
\end{table}

 We use PyTorch\cite{paszke2017automatic} for the network implementation. In the proposed multi-label classification RBM we use 60 seconds of aggregated load sampled at 1Hz as input to the model. For hidden unit following sizes are tried $32$, $64$, $128$, and $256$, we find $128$ to be best. The learning rate is set to $0.001$ for all our experiments. We use $k=2$ steps of sampling in CD \cite{CD2002hinton} algorithm to train our model. For inference, we apply sigmoid activation to the output of our model and threshold at $0.5$.

Macro F1 and Micro F1 scores are the two metrics which are commonly used to evaluate the performance of Multi-label classifiers. Appliance-level energy error is computed for each device to evaluate disaggregation performance. Macro F1 score is average of individual F1 score of all the classes so it could be biased towards a class with fewer samples. The Micro F1 score indicates the overall performance of the classifier. It is computed by stacking up samples from all the classes. The F1 score of an individual class is given by \autoref{eqn 13},

\begin{equation}
\label{eqn 13}
F1=\frac{{2 \times TP}}{{2 \times TP + FN + FP}}
\end{equation}
Where TP is the number of true positives, FN is the number of false negatives and FP is the number of false positives.

The appliance-level error also known as Normalized energy error (NEE) is a standard metric which is used in almost every prior study in this area and it is given as \autoref{eqn 14},

\begin{equation}
\label{eqn 14}
NEE = \frac{{\sum\limits_t {|P_t^n - \hat P_t^n|} }}{{\sum\limits_t {P_t^n} }}
\end{equation}
where $P_t^n$ is the power consumption of the appliance \textit{n} at any time instant \textit{t}.

\autoref{tab1} and \autoref{tab2} present the F1-Score and correspondingly obtained disaggregation error for each target device in both the datasets. \autoref{tab3} and \autoref{tab4} contain micro and macro F1-Scores yielded by the state-of-the-art and proposed algorithm on the REDD and Pecan Street dataset respectively. Our proposed model yields the best results regarding classification measures and gives comparable disaggregation accuracy. Although best classification accuracy should reflect the least disaggregation error, here it is not so. This mismatch engenders an ambiguity in results.

We would like to clarify it with an example. Suppose true labels for two hours of aggregate consumption of four devices are 1 0 0 1 and  0 1 1 0 whereas the predicted labels are  0 1 1 0 and 1 0 0 1 respectively. For the given case F1-Score would be zero as all the identified states are wrong. For the same case, disaggregation accuracy would be 100 \% as the number of identified active appliances exactly matches the number of true active appliances. This example explains why techniques, such as LC-DDL, gives the best disaggregation accuracy but worst F1-Scores. Therefore in such a framework, the performance of an algorithm should be judged only after looking at both metrics collectively.

\begin{table}
\begin{center}
\caption {\textbf{Performance Evaluation on Pecan Street}}
\label{tab4}
\begin{tabular}{c c c c}
\toprule[0.2mm]
\textbf{Method}    & \textbf{Macro F1-Score} &\textbf{ Micro F1-Score} \\
    \midrule
     MLkNN              &  0.6183          &  0.6194                    \\

     RAkEL              &  0.5872          &  0.6019                     \\

     LC-DDL             &  0.5214          &  0.5332                   \\

     MLC-RBM        &\textbf{0.7080}   &\textbf{0.7123}          \\
    \bottomrule[0.2mm]
\end{tabular}
\end{center}
\end{table}

\section{Conclusion}
This work proposes a new technique for NILM framed as a multi-label classification problem. The proposed multi-label classification RBM has good reconstruction ability and when combined with multi-label supervision also provides good classification accuracy. This technique does not require any appliance-level data which makes the task completely non-intrusive. We compare the proposed technique with all the prior works where NILM was transformed as a multi-label classification task. We have performed an experimental evaluation of the proposed work on two widely used datasets. Our proposed model yields the best results in term of classification accuracy and comparable results regarding energy disaggregation. Although we have used multi-label RBM for NILM, it is a generic approach and can be used for solving any multi-label classification problem. In the future, we plan to benchmark it against existing algorithms on established multi-label classification datasets.

\bibliographystyle{IEEEbib}
\bibliography{ms}

\end{document}